\documentclass{article}
\usepackage[utf8]{inputenc}
\usepackage{main}
\usepackage{microtype}
\usepackage{subcaption}
\usepackage{graphicx}
\usepackage{times}
\usepackage{latexsym}
\usepackage{amsmath}
\usepackage{float}
\usepackage{footnote}
\usepackage{enumitem}
\usepackage{bm}
\usepackage{arydshln}
\usepackage{booktabs}
\usepackage{multicol}
\usepackage{multirow}
\usepackage{color}
\usepackage{xcolor}     
\usepackage{colortbl}
\usepackage{bbding}
\usepackage{makecell}
\usepackage{mathtools}
\usepackage{imakeidx}
\usepackage{longtable}
\usepackage{wrapfig}
\makeindex
\usepackage{arydshln}
\usepackage{lipsum}
\usepackage{natbib}
\usepackage[toc]{multitoc}
\usepackage[edges]{forest}
\usepackage[normalem]{ulem}
\definecolor{mydarkblue}{rgb}{0,0.08,0.45}
\usepackage[colorlinks=true,linkcolor=mydarkblue,citecolor=mydarkblue,filecolor=mydarkblue,urlcolor=mydarkblue]{hyperref}
\usepackage{CJKutf8}
\usepackage{awesomebox} 
\usepackage{bbding}
\usepackage[most]{tcolorbox}
\usepackage{booktabs}
\usepackage{geometry}
\definecolor{wkblue}{rgb}{0.2, 0.3, 0.6}
\definecolor{meta-color}{rgb}{0.5, 0.5, 0.5}
\usepackage{amsmath}
\usepackage{enumitem}

\usepackage[tikz]{bclogo}
\usepackage[framemethod=tikz]{mdframed}
\definecolor{bgblue}{RGB}{245,243,253}
\definecolor{ttblue}{RGB}{91,194,224}

\mdfdefinestyle{mystyle}{%
  rightline=true,
  innerleftmargin=10,
  innerrightmargin=10,
  outerlinewidth=3pt,
  topline=false,
  rightline=true,
  bottomline=false,
  skipabove=\topsep,
  skipbelow=\topsep
}

\newtcolorbox{myboxi}[1][]{
  breakable,
  title=#1,
  colback=red!5,
  colbacktitle=red!5,
  coltitle=black,
  fonttitle=\bfseries,
  bottomrule=0pt,
  toprule=0pt,
  leftrule=2pt,
  rightrule=2pt,
  titlerule=0pt,
  arc=0pt,
  outer arc=0pt,
  colframe=red,
}

\newtcolorbox{myboxnote}[1][]{
  breakable,
  title=#1,
  colback=orange!0,
  colbacktitle=orange!0,
  coltitle=black,
  fonttitle=\bfseries,
  bottomrule=0pt,
  toprule=0pt,
  leftrule=2pt,
  rightrule=2pt,
  titlerule=0pt,
  arc=0pt,
  outer arc=0pt,
  colframe=orange,
}

\newtcolorbox{myboxii}[1][]{
  breakable,
  freelance,
  title=#1,
  colback=white,
  colbacktitle=white,
  coltitle=black,
  fonttitle=\bfseries,
  bottomrule=0pt,
  boxrule=0pt,
  colframe=white,
  overlay unbroken and first={
  \draw[red!75!black,line width=3pt]
    ([xshift=5pt]frame.north west) -- 
    (frame.north west) -- 
    (frame.south west);
  \draw[red!75!black,line width=3pt]
    ([xshift=-5pt]frame.north east) -- 
    (frame.north east) -- 
    (frame.south east);
  },
  overlay unbroken app={
  \draw[red!75!black,line width=3pt,line cap=rect]
    (frame.south west) -- 
    ([xshift=5pt]frame.south west);
  \draw[red!75!black,line width=3pt,line cap=rect]
    (frame.south east) -- 
    ([xshift=-5pt]frame.south east);
  },
  overlay middle and last={
  \draw[red!75!black,line width=3pt]
    (frame.north west) -- 
    (frame.south west);
  \draw[red!75!black,line width=3pt]
    (frame.north east) -- 
    (frame.south east);
  },
  overlay last app={
  \draw[red!75!black,line width=3pt,line cap=rect]
    (frame.south west) --
    ([xshift=5pt]frame.south west);
  \draw[red!75!black,line width=3pt,line cap=rect]
    (frame.south east) --
    ([xshift=-5pt]frame.south east);
  },
}

\usepackage{fancyhdr} 
\usepackage{blindtext} 

\DeclareCaptionFont{black}{\color{black}}

\definecolor{myblue}{rgb}{0.9, 0.1, 0.94}
\definecolor{mygreen}{rgb}{0.64, 0.56, 0.88}
\definecolor{myyellow}{rgb}{0.68, 0.6, 0.1}
\definecolor{fancygreen}{rgb}{0.33, 0.68, 0.20}
\definecolor{salmon}{rgb}{0.94, 0.52, 0.49}
\definecolor{tablegreen}{rgb}{0.82, 0.94, 0.75}
\definecolor{tableblue}{rgb}{0.81, 0.90, 0.94}
\definecolor{tablered}{rgb}{0.97, 0.85, 0.85}
\definecolor{tableorange}{rgb}{0.96, 0.85, 0.81}

\newenvironment{itemize*}%
 {\leftmargini=10pt\begin{itemize}%
  \setlength{\itemsep}{0pt}%
  \setlength{\parskip}{0pt}%
  }%
 {\end{itemize}}
\newenvironment{enumerate*}%
 {\begin{enumerate}%
  \setlength{\itemsep}{0pt}%
  \setlength{\parskip}{0pt}}%
 {\end{enumerate}}

\usepackage{xcolor}
\usepackage{listings}

\newcommand\JSONnumbervaluestyle{\color{blue}}
\newcommand\JSONstringvaluestyle{\color{red}}

\newif\ifcolonfoundonthisline

\makeatletter

\lstdefinestyle{json}
{
  showstringspaces    = false,
  keywords            = {false,true},
  alsoletter          = 0123456789.,
  morestring          = [s]{"}{"},
  stringstyle         = \ifcolonfoundonthisline\JSONstringvaluestyle\fi,
  MoreSelectCharTable =%
    \lst@DefSaveDef{`:}\colon@json{\processColon@json},
  basicstyle          = \ttfamily,
  keywordstyle        = \ttfamily\bfseries,
}

\newcommand\processColon@json{%
  \colon@json%
  \ifnum\lst@mode=\lst@Pmode%
    \global\colonfoundonthislinetrue%
  \fi
}

\lst@AddToHook{Output}{%
  \ifcolonfoundonthisline%
    \ifnum\lst@mode=\lst@Pmode%
      \def\lst@thestyle{\JSONnumbervaluestyle}%
    \fi
  \fi
  \lsthk@DetectKeywords%
}

\lst@AddToHook{EOL}%
  {\global\colonfoundonthislinefalse}

\makeatother

\usepackage{etoolbox}
\usepackage{natbib}
\usepackage{url}
\newcounter{bibcount}
\makeatletter
\patchcmd{\@lbibitem}{\item[}{\item[\hfil\stepcounter{bibcount}{[\thebibcount]}}{}{}
\renewcommand\NAT@bibsetup%
  [1]{\setlength{\leftmargin}{\bibhang}\setlength{\itemindent}{-\parindent}%
      \setlength{\itemsep}{\bibsep}\setlength{\parsep}{\z@}}
\makeatother

\newcommand*\samethanks[1][\value{footnote}]{\footnotemark[#1]}

\author{%
Yiwei Qin$^{1, 4}$\thanks{~~Co-first authors}\space\space\space\space Xuefeng Li$^{1, 4}$\samethanks\space\space\space\space Haoyang Zou$^{4}$\samethanks\space\space\space\space Yixiu Liu$^{1, 4}$\samethanks\space\space\space\space Shijie Xia$^{1, 4}$\samethanks\\
\textbf{Zhen Huang$^{4}$\space\space\space Yixin Ye$^{1, 4}$\space\space\space Weizhe Yuan$^{2}$\space\space\space Hector Liu$^{3}$\space\space\space Yuanzhi Li$^{3}$\space\space\space Pengfei Liu$^{1, 4}$}\thanks{~~Corresponding author}\\
$^1$Shanghai Jiao Tong University, $^2$New York University,\\ $^3$MBZUAI, $^4$Generative AI Research Lab (GAIR)}

\begin{document}

\title{O1 Replication Journey: A Strategic Progress Report -- Part 1}

\maketitle
\thispagestyle{fancy}
\fancyhead{}
\lhead{\includegraphics[height=0.67cm]{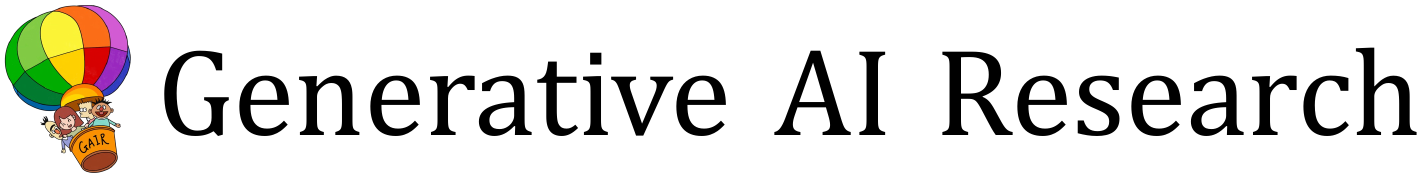}}
\renewcommand{\headrulewidth}{0pt}
\setlength{\headsep}{0mm}

\begin{abstract}

This paper introduces a pioneering approach to artificial intelligence research, embodied in our O1 Replication Journey. In response to the announcement of OpenAI's groundbreaking O1 model, we embark on a \textit{transparent, real-time exploration} to replicate its capabilities while \textit{reimagining} the process of conducting and communicating AI research. Our methodology addresses critical challenges in modern AI research, including the insularity of prolonged team-based projects, delayed information sharing, and the lack of recognition for diverse contributions. By providing comprehensive, real-time documentation of our replication efforts, including both successes and failures, we aim to foster open science, accelerate collective advancement, and \textbf{lay the groundwork for AI-driven scientific discovery}.
Our research progress report diverges significantly from traditional research papers, offering continuous updates, full process transparency, and active community engagement throughout the research journey. 
Technologically, we proposed the ``\textbf{journey learning}'' paradigm, which encourages models to learn not just shortcuts, but the complete exploration process, including trial and error, reflection, and backtracking. With only \textbf{327 training samples} and without any additional tricks, journey learning outperformed conventional supervised learning by \textbf{over 8\%} on the MATH dataset, demonstrating its extremely powerful potential. We believe this to be the most crucial component of O1 technology that we have successfully decoded.
We share valuable resources including technical hypotheses and insights, cognitive exploration maps, custom-developed tools, etc at \url{https://github.com/GAIR-NLP/O1-Journey}.

\end{abstract}

\begin{figure}[ht]
    \centering
    \includegraphics[width=1.0\linewidth]{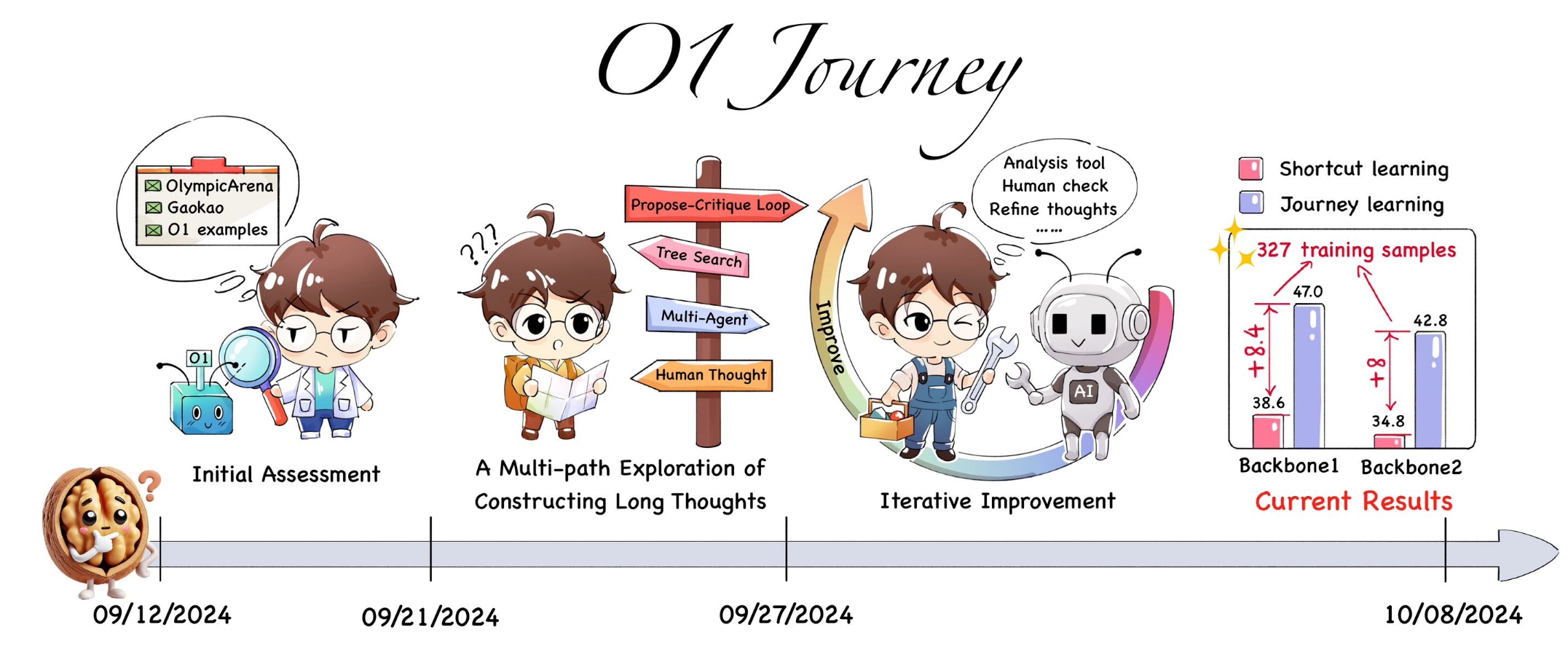}
    \caption{
     Illustration of our O1 replication journey from September 12 to October 8, 2024. It depicts four key stages: Initial Assessment, Multi-path Exploration, Iterative Improvement, and Current Results. The journey culminates in our novel ``\textbf{journey learning}'' approach, which significantly outperforms traditional ``shortcut learning'' methods. With only 327 training samples, our journey learning technique surpassed shortcut learning by 8.4\% and 8.0\% respectively on the MATH500~\citep{DBLP:conf/iclr/LightmanKBEBLLS24}. \includegraphics[scale=0.06]{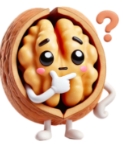} denotes our \href{https://gair-nlp.github.io/walnut-plan}{Walnut Plan}, which aims to revolutionize AI by developing systems capable of deep scientific thinking, ultimately enabling AI-driven breakthroughs in human knowledge and discovery.
    }
    \label{fig:enter-label}
\end{figure}

\newpage

\pagestyle{fancy}
\lhead{\rightmark}
\renewcommand{\headrulewidth}{0.7pt}
\setlength{\headsep}{5mm}


\clearpage

\section{Chronological Overview of the O1 Exploration Journey}
\begin{figure*}[!ht]
    \centering
    \centering
    \includegraphics[width=0.99\linewidth]{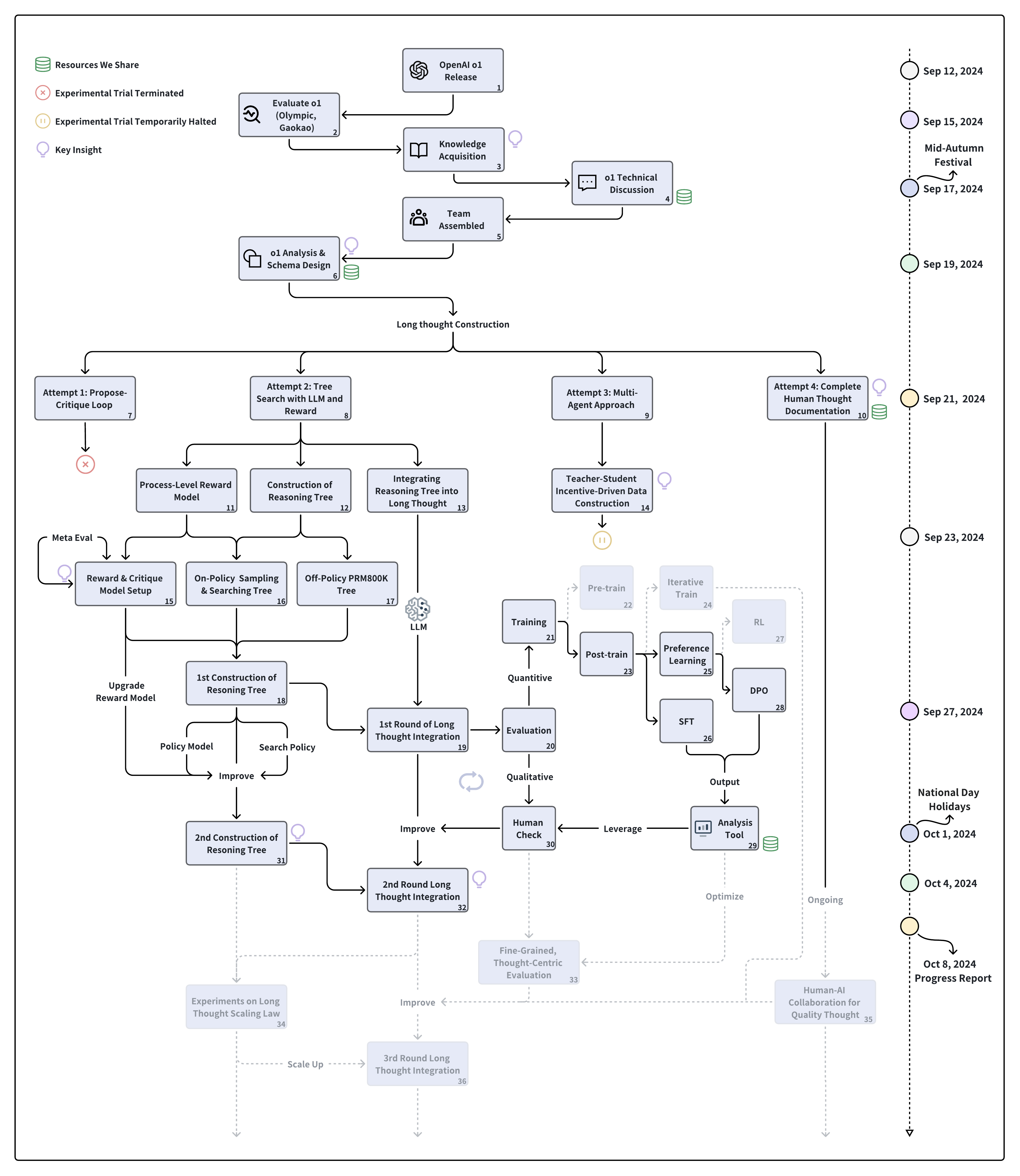} 
    \caption{
    This figure outlines our research journey exploring OpenAI's o1 technology from its release through October 8, 2024. A timeline tracks our progress chronologically, with research activities flowing vertically in the main diagram. Following the o1 release, we progressed from initial evaluation and knowledge acquisition to team assembly and analysis. Our exploration then focused on four long thought construction attempts. The second attempt, our core exploration, splits into three tracks: Process-Level Reward Model, Construction of Reasoning Tree, and Integrating Reasoning Tree into Long Thought (detailed explanations of specific nodes can be found in Table \ref{details_table}). These converge in an iterative cycle of model improvement, including both quantitative and qualitative evaluation. The diagram's right side illustrates our training pipeline, featuring pre-training, iterative training, and optimization techniques. Solid black elements represent completed paths and milestones, while gray dashed elements indicate planned future explorations. This visualization captures both our achievements and future research directions in o1 technology development.
}
    \label{research_journey}
\end{figure*}

\newpage

\section{Introduction}



The landscape of artificial intelligence research has been dramatically altered by the announcement of OpenAI's O1 model, a purportedly groundbreaking language model capable of complex reasoning tasks. Despite the excitement generated by this announcement, the AI community finds itself in a peculiar position: we know of O1's existence and its claimed capabilities, but the details of its implementation, training data, and even its complete outputs remain shrouded in mystery. This lack of transparency not only hampers technological progress but also raises important questions about the open nature of scientific advancement in the AI field.
It is within this context that our team embarked on the O1 Replication Journey. Our primary goal is \textbf{not to achieve performance parity with OpenAI's O1} - a task we acknowledge as extremely challenging given the limited information and resources available. Instead, our mission is to \textbf{transparently document} and share our exploration process, focusing on the fundamental questions we encounter, uncovering new scientific questions, and sharing our trial-and-error experiences with the broader AI community. By doing so, we aim to \textbf{reduce the total collective cost of trial-and-error in the world} and identify the key factors contributing to O1's reported success.

This report's structure marks a significant departure from traditional scientific publications, addressing key challenges in modern AI research. In an era of prolonged, team-based AI projects, we aim to combat information isolation and researcher burnout through enhanced transparency and real-time feedback. Additionally, this report represents a bold reimagining of AI research methodology. It aims not only to provide a valuable reference for current O1 replication efforts but also to establish a new paradigm for future AI research and broader scientific exploration. Through this innovative approach, we strive not only to achieve technological breakthroughs but also to actively shape a more open, collaborative, and responsible scientific culture. Simultaneously, we are \textbf{accumulating invaluable learning materials for future AI systems capable of scientific discovery}, thus laying the groundwork for the next generation of artificial intelligence in scientific research.

\begin{figure*}[!ht]
    \centering
    \centering
    \includegraphics[width=0.99\linewidth]{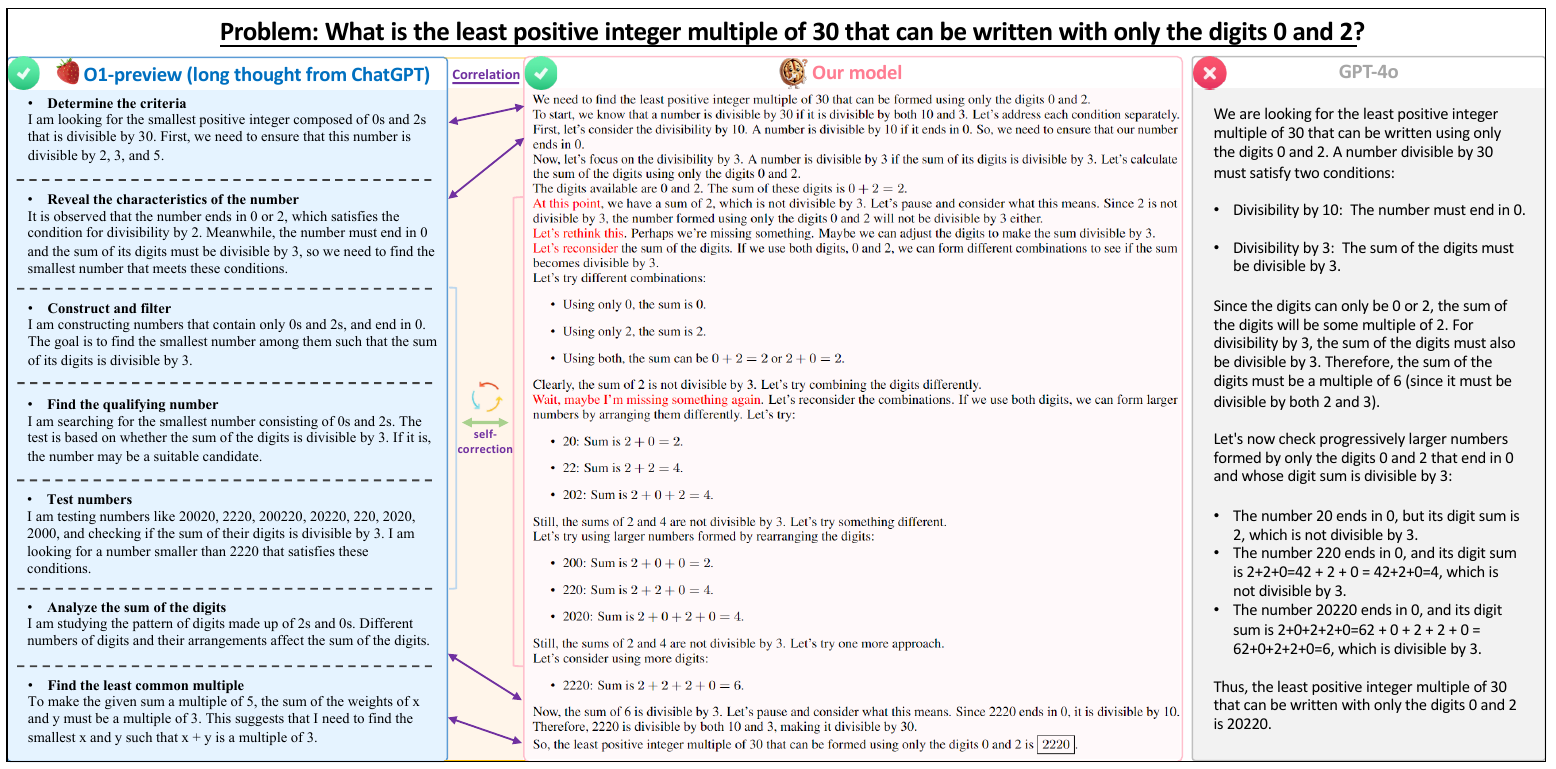} 
    \caption{A case study comparing our model with OpenAI O1-preview and GPT-4o in solving math problems.}
    \label{case_compare}
\end{figure*}

Our initial explorations have already yielded intriguing insights into the potential mechanisms behind O1's reported capabilities. A key breakthrough in our research at the current stage is the proposed ``\textbf{journey learning}'' paradigm, which represents a fundamental shift in how we approach model training. This innovative method encourages models to learn not just shortcuts to solutions, but the complete exploration process, including trial and error, reflection, and backtracking (see Figure~\ref{case_compare}). The power of this approach is evident in its performance: \textbf{with only 327 training samples and without any additional tricks, journey learning outperformed conventional supervised learning by over 8\% on the MATH dataset}, demonstrating its extremely powerful potential. We believe this to be the most crucial component of O1 technology that we have successfully decoded so far.

Through this journey, we anticipate a multifaceted impact on the field of AI research and development: (1) we expect to gain deeper insights into the fundamental principles underlying advanced language models, potentially uncovering key mechanisms that contribute to O1's reported capabilities.
(2) Moreover, by championing transparency and real-time sharing of our findings, we aim to foster a more open and collaborative AI research ecosystem, encouraging knowledge exchange and collective problem-solving. 
(3) Finally, by meticulously documenting our entire journey, including both successes and failures, we will create an invaluable dataset for training future AI systems in scientific discovery, laying the groundwork for the next generation of AI-driven research methodologies.

As part of our commitment to open science, we will be releasing numerous valuable resources throughout this journey. These include:

(1) Our detailed hypotheses regarding the O1 technical stack, along with a comprehensive map of our cognitive exploration path. This resource provides insight into our strategic thinking and decision-making processes throughout the replication attempt.
(2) A collection of insights and positive outcomes derived from our trial-and-error experiences. This compilation offers valuable lessons learned and unexpected discoveries that may benefit the broader AI research community.
(3) Extensive documentation of our cognitive processes, including discussion presentations and brainstorming sessions. These materials offer a transparent look into our team's collaborative problem-solving approach and idea generation.
(4) Preliminary results and experimental data from our initial efforts, as well as access to our custom-developed annotation platform. These resources showcase our early progress and provide practical tools for researchers engaged in similar endeavors.


\section{Why We Created Progress Report?}

\begin{table*}[!t]
    \centering
    \setlength{\tabcolsep}{6pt}
    {
    \fontsize{10}{13}\selectfont 
    \renewcommand\arraystretch{1.2}
    \resizebox{\linewidth}{!}{
        \begin{tabular}[l]{llll}
        \toprule
        \textbf{Aspects} &  & \textbf{Traditional Research Paper} & \textbf{Proposed Progress Report} \\ \midrule
        \multirow{3}{*}{\textbf{Research Process}} 
        & Timing of Publication & Published after research completion & Real-time updates throughout the research process \\
        & Length of Research & Suitable for shorter-term projects & Designed for prolonged, team-based endeavors \\
        & Handling of Setbacks & Typically not reported in detail & Candidly shared as valuable learning experiences \\ \midrule
        \multirow{3}{*}{\textbf{Information Sharing}} 
        & Information Flow & Limited until publication & Continuous sharing of insights and findings \\
        & Transparency & Focus on successful outcomes & Full disclosure of process, including failures \\
        & Data Sharing & Often limited to final results & Includes interim data, tools, and methodologies \\ \midrule
        \multirow{3}{*}{\textbf{Impact and Value}} 
        & Impact on Motivation & Delayed gratification until publication & Ongoing feedback and recognition \\
        & Reproducibility & Often challenging due to limited details & Enhanced through comprehensive documentation \\
        & AI Training Potential & Limited to final outcomes & Rich dataset including full exploration process \\
        \bottomrule
        \end{tabular}
    }
    \caption{Comparison between ``Traditional Research Paper'' and ``Proposed Progress Report''.}
    \label{tab:comparison}
    }
\end{table*}

In the rapidly evolving landscape of artificial intelligence research, traditional methodologies and reporting practices are increasingly proving inadequate to address the complexities and scale of modern AI projects. This report represents a pioneering effort to reimagine the process of conducting and communicating AI research. By providing a comprehensive, real-time account of our journey to replicate the groundbreaking O1 model, we aim to address critical challenges in contemporary AI research, foster open science, redefine scientific communication, lay the groundwork for AI-driven scientific discovery, and promote responsible AI development. \textbf{What follows is not merely a documentation of our findings, but a bold proposition for a new paradigm in scientific exploration and collaboration in the AI era.}

\begin{enumerate}
    \item \textbf{Addressing the Challenges of Modern AI Research}: The rapid evolution of artificial intelligence technologies has ushered in a new era of research paradigms, characterized by \textbf{prolonged, team-based endeavors} that often span six months or more. This shift, while conducive to breakthrough innovations, has inadvertently introduced novel challenges to the scientific process. The inherent insularity of extended team collaborations frequently \textbf{results in a diminished flow of information to the broader scientific community}. Moreover, the protracted nature of these projects often leads to \textbf{delayed gratification for researchers, potentially fostering anxiety and diminished motivation throughout the research journey}. Additionally, the complexity of large-scale team projects complicates the recognition of individual contributions, potentially eroding the traditional academic incentive structures. Our progress report methodology \textbf{aims to address these emergent challenges by enhancing transparency, facilitating real-time feedback and recognition, and encouraging sustained commitment to long-term research initiatives}.
    \item \textbf{Fostering Open Science and Collective Advancement}: In the spirit of open science and collective advancement, \textbf{the primary impetus behind this report is to disseminate the invaluable insights, resources, and lessons gleaned from our endeavor to replicate the O1 model}. This approach transcends the mere sharing of a trained model; it encompasses a comprehensive documentation of the tools, datasets, and methodologies employed throughout our exploratory process. By candidly sharing our setbacks and unsuccessful attempts, we aim to provide educational value that often surpasses that of mere success stories. This transparency is intended to assist other researchers in navigating potential pitfalls, thereby accelerating progress across the field. Furthermore, by elucidating our thought processes and innovative approaches, we aspire to catalyze creativity within the community, fostering the generation of novel ideas and methodologies.
    \item \textbf{Laying the Foundation for AI in Scientific Discovery}: The meticulous documentation of our scientific exploration process holds profound significance, particularly in the context of rapidly advancing AI capabilities. \textbf{By recording our exploration process in its entirety, including both successes and failures, we are cultivating a unique and invaluable dataset. This comprehensive record is crucial for training AI models that genuinely comprehend scientific methodologies, mirroring the approach validated by the O1 model. The success of O1 underscores the importance of AI systems learning not just outcomes, but the complete scientific exploration process, including trial and error}. Our report captures not only technical details but also decision rationales, sources of inspiration, and thought processes. These "human factors" are essential for training AI models capable of authentic scientific discovery. Moreover, this approach has interdisciplinary value, offering a template for research documentation and knowledge sharing that can foster innovation across various scientific domains.
    \item \textbf{Promoting Responsible AI Development}: 
    In our pursuit of technological breakthroughs, we remain acutely aware of the potential societal impacts and ethical considerations associated with AI development. Through detailed documentation of our research processes and decision-making, we \textbf{establish a high standard of transparency}, which is crucial for cultivating public trust in AI research. Our report goes beyond technical specifics, incorporating ongoing discussions and reflections on potential societal impacts, thereby demonstrating the integration of ethical considerations throughout the technological development process. This holistic approach contributes to the cultivation of a more responsible and ethically-minded AI research culture.and recognition, and encouraging sustained commitment to long-term research initiatives.
\end{enumerate}


\section{Journey Learning: A New Paradigm Shift from ``Shortcut Learning''} \label{sec:journey-learning}

\begin{figure}[!ht]
    \centering
    \begin{subfigure}{0.25\textwidth}
        \centering
        \includegraphics[width=\textwidth]{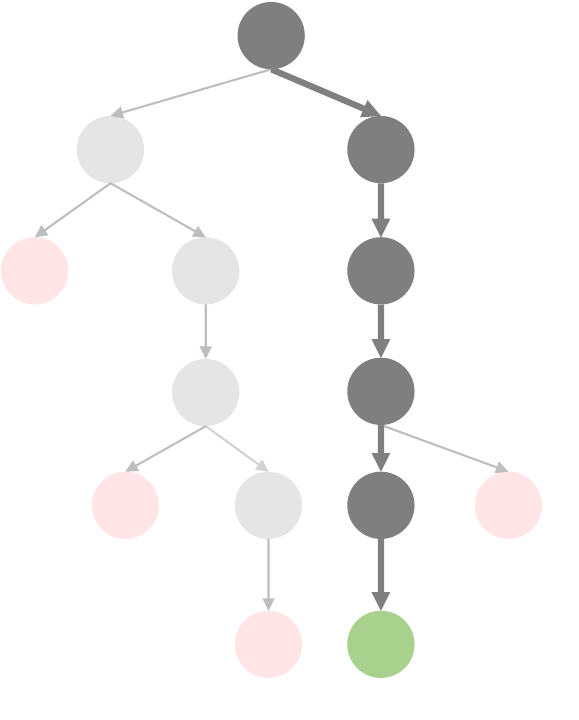} 
        \caption{Shortcut learning.}
        \label{shortcut}
    \end{subfigure}
    \begin{subfigure}{0.25\textwidth}
        \centering
        \includegraphics[width=\textwidth]{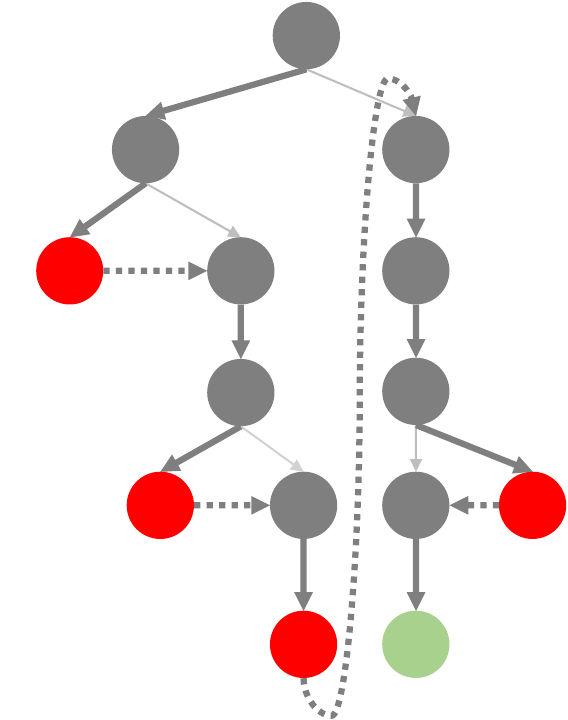}
        \caption{Journey learning}
        \label{journey}
    \end{subfigure}
     \hspace{0.07\textwidth}  
    \begin{subfigure}{0.35\textwidth}
        \centering
        \includegraphics[width=\textwidth]{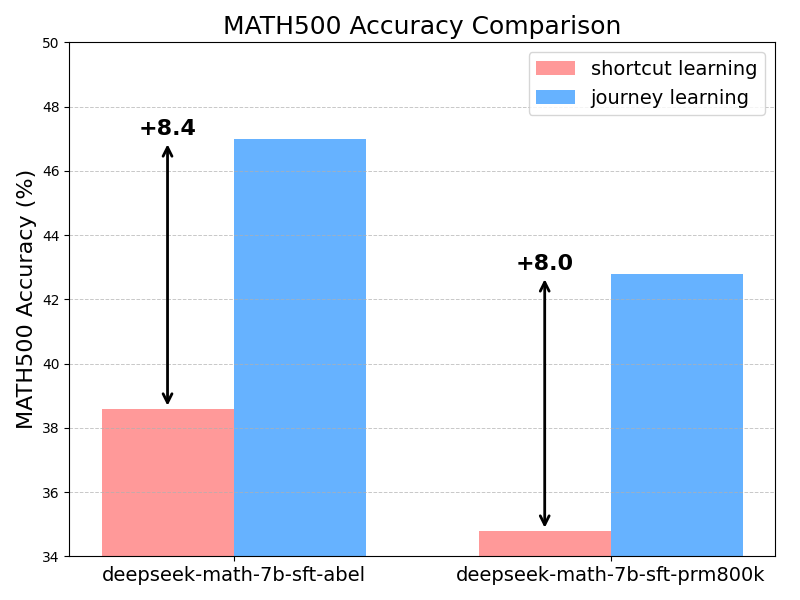}
        \caption{Performance Comparison}
        \label{journey}
    \end{subfigure}
    \caption{A paradigm shift from ``shortcut learning'' to ``journey learning''. 
    A searching tree for reasoning tasks. For the math problem-solving task, the root node represents the initial problem, while the leaf nodes are final conclusions. Green nodes indicate correct answers, and red nodes incorrect ones.
    Traditionally, learning focused on supervised training of a direct root-to-leaf shortcut path. This work, however, explores supervised learning of the entire exploration path, encompassing trial-and-error and correction processes.
    (c) Performance of ``shortcut learning'' and ``journey learning'' on MATH500~\citep{DBLP:conf/iclr/LightmanKBEBLLS24}. The base models are deepseek-math-7b-base fine-tuned on Abel training data and PRM800K separately.
    }
    \label{shor_long-comparison}
\end{figure}


We claim that most existing approaches to machine learning or large language model training (e.g., supervised fine-tuning) can be characterized as ``\textbf{shortcut learning}.'' This traditional paradigm, while potentially effective in specific, well-defined tasks, shows significant limitations when faced with complex, dynamic, and open-ended problems. Shortcut learning is defined by several key characteristics:
(I) Quick results orientation: It emphasizes achieving specific performance metrics or completing particular tasks in a short time frame. (2) Heavy data dependency: Performance improvements often rely on increasing the volume of training data rather than enhancing the learning algorithms themselves. (3) Limited generalization: Performance can deteriorate dramatically in scenarios outside the distribution of the training data. (4) Lack of self-correction: These systems typically lack the ability to identify and correct their own errors.
While shortcut learning has driven many advances in AI, it struggles to produce \textit{truly intelligent} and reliable AI systems capable of handling the complexities of real-world challenges. As we pursue more advanced forms of artificial intelligence or even \emph{superintelligence}, the limitations of this approach become increasingly apparent.
Recognizing these shortcomings, we propose a new paradigm called ``\textbf{journey learning}." This innovative approach represents \textbf{more than just a learning method}; it's a new paradigm for AI development. Journey learning is designed to enable AI systems to progress continuously through \emph{learning, reflection, backtracking, and adaptation}, much like humans do, thereby exhibiting higher levels of intelligence.

Journey learning represents a significant advancement over shortcut learning. While shortcut learning often falters in complex, dynamic environments, journey learning is specifically designed to thrive in such scenarios. It aims to create AI systems that are not just narrow, task-specific tools, but adaptable, reasoning entities capable of handling the nuances and complexities of real-world challenges.
This new paradigm holds the promise of more capable, adaptable, and human-like AI that can better serve and interact with humans across various domains. As we continue to develop and refine the journey learning paradigm, we expect it to open new possibilities in AI research and applications, potentially revolutionizing our approach to artificial intelligence and its role in our society.


\begin{table*}[!t]
    \centering
    \setlength{\tabcolsep}{6pt}
    {
    \fontsize{10}{13}\selectfont 
    \renewcommand\arraystretch{1.2}
    \resizebox{\linewidth}{!}{
        \begin{tabular}[l]{lll}
        \toprule
        \textbf{Characteristic} & \textbf{Shortcut Learning} & \textbf{Journey Learning} \\ \midrule
        \multirow{1}{*}{\textbf{Learning Depth}} & Surface features and simple correlations & Deep causal relationships and underlying principles \\ 
        \multirow{1}{*}{\textbf{Reasoning Ability}} & Limited, struggles with complex reasoning & Powerful, demonstrates human-like reasoning \\ 
        \multirow{1}{*}{\textbf{Self-Improvement}} & Lacks self-correction mechanisms & Continuous self-assessment and improvement \\ 
        \multirow{1}{*}{\textbf{Generalization}} & Limited, easily affected by data distribution changes & Strong, can handle new situations \\ 
        \multirow{1}{*}{\textbf{Innovation Capacity}} & Limited, struggles to solve new problems & High, can generate innovative solutions \\ 
        \multirow{1}{*}{\textbf{Data Dependency}} & Highly dependent on large training datasets & More focused on quality and learning strategies \\ 
        \multirow{1}{*}{\textbf{Interpretability}} & Poor, often seen as a ``black box'' & Better, can track internal reasoning processes \\ 
        \multirow{1}{*}{\textbf{Ethical Considerations}} & May unintentionally amplify data biases & Easier to implement ethical constraints and adjustments \\ 
        \multirow{1}{*}{\textbf{Security}} & Vulnerable to adversarial attacks & More robust, able to identify potential threats \\ 
        \multirow{1}{*}{\textbf{Long-term Value}} & Quick results in specific tasks & Paves the way for AGI development \\ 
        \multirow{1}{*}{\textbf{Human Analogy}} & Exam-oriented education, crash courses & Comprehensive education, lifelong learning \\ 
        \bottomrule
        \end{tabular}
    }
    \caption{Comparison between Shortcut Learning and Journey Learning.}
    \label{tab:comparison}
    }
\end{table*}

\section{Background}

\paragraph{Process-level Reward Model}
Process reward models (PRMs) are used to provide fine-grained evaluations of responses from LLMs~\citep{DBLP:conf/iclr/LightmanKBEBLLS24,uesato2022solving,xia2024evaluating}, especially in the area of mathematical reasoning. By accurately assessing the correctness of each step, PRMs can enhance post-training quality~\citep{wang2024math, sun2024easy} and improve accuracy during inference through various search methods~\citep{luo2024improve, wang2024q}. Implementing PRMs can involve using proprietary models with advanced prompting techniques~\citep{hao2024llm} or training with step-level supervision data~\citep{xia2024evaluating, wang2024math}. The latter approach is challenging because it requires high-quality annotated data~\citep{xia2024evaluating}. This has led to interest in using reinforcement learning principles, which model the multi-step reasoning process as a Markov Decision Process (MDP) and use techniques like Monte Carlo Tree Search~\citep{silver2016mastering} to estimate the value of each step, either online~\citep{chen2024alphamath} or offline~\citep{wang2024math}.

%

\paragraph{COT Theory}
Chain-of-thought (CoT) prompting has significantly advanced the reasoning capabilities of LLMs. Foundational studies demonstrate that providing intermediate reasoning steps enhances performance on complex tasks, such as arithmetic and commonsense reasoning~\citep{wei2022chain}. Additionally, theoretical investigations reveal that CoT empowers decoder-only transformers by enabling inherently serial computation, which is otherwise lacking, particularly in low-depth transformers~\citep{li2024chain}. Recent studies also reveal CoT prompting enhances LLMs by showing that even constant-sized autoregressive Transformers can solve complex tasks like arithmetic and decision-making through CoT derivations, using circuit complexity theory~\citep{feng2024towards}. Recent work emphasizes the integration of ``error-correction'' data into the pretraining stage to enhance reasoning accuracy, showing that such data can lead to higher accuracy without the need for multi-round prompting~\citep{ye2024physicslanguagemodels22}. Overall, these findings underscore the pivotal role of CoT prompting in enhancing LLM performance and accessibility in complex reasoning tasks.

\paragraph{Internal Thought}
The exploration of internal thought in AI models has evolved as researchers emphasize the need for models to reflect on their reasoning and refine their outputs. Early work like STaR~\citep{zelikman2022star}  proposed bootstrapping reasoning by having models generate rationales that explain their decisions, allowing them to improve their performance on complex tasks through iterative refinement. Building on this, Quiet-STaR~\citep{zelikman2024quiet}  generalizes the approach by training language models to generate rationales after each token, helping them predict and explain future text more effectively. \citet{zhang2024learn} further expanded this line of work by embedding reflection within each training instance, encouraging models to review their decisions and consider alternative reasoning paths. RISE~\citep{qu2024recursive} introduced a method for recursive introspection, where models iteratively adjust their responses after detecting errors, aiming for self-improvement over multiple attempts. These developments illustrate the growing focus on enabling AI models to engage in reflective, self-correcting processes, enhancing their ability to handle more complex reasoning tasks.

\paragraph{Inference Time Scaling}
Recent studies have demonstrated that scaling inference time can provide more efficient improvements in model performance~\cite{Sardana2023BeyondCA, Snell2024ScalingLT} compared to traditional scaling approaches such as increasing model parameters or training data volume. While parameter scaling has been a dominant paradigm in advancing model capabilities~\cite{kaplan2020scalinglawsneurallanguage, brown2020languagemodelsfewshotlearners, chowdhery2022palmscalinglanguagemodeling}, it often leads to diminishing returns and significant computational overhead. In contrast, allowing models more time to process and refine their outputs during inference has emerged as a promising alternative scaling dimension~\cite{madaan2023selfrefineiterativerefinementselffeedback}.
Inference time scaling offers several advantages: 1) Resource efficiency, utilizing existing model capacities more thoroughly; 2) Adaptable computation, allocating more processing time to complex tasks; and 3) Improved reasoning through step-by-step problem solving or iterative refinement~\cite{yao2023treethoughtsdeliberateproblem, Cobbe2021TrainingVT}. Empirical evidence suggests that doubling inference time often yields performance improvements comparable to those achieved by significantly increasing model size, but at a fraction of the computational cost~\cite{Snell2024ScalingLT}.
Some successful implementations include deliberation mechanisms and iterative refinement protocols~\cite{Huang2022LargeLM, Miao2023SelfCheckUL}, which have shown particular promise in tasks requiring complex reasoning or creative generation.

\paragraph{Search-to-thought}
In recent years, the shift from traditional search-based methods to implicit reasoning approaches has significantly advanced AI research~\cite{ruoss2024grandmaster}. Classic systems like Deep Blue~\cite{CAMPBELL200257} relied heavily on explicit search algorithms such as alpha-beta pruning and Monte Carlo Tree Search to achieve superhuman performance~\cite{silver2017mastering}. However, with the advent of deep learning, Chain of Thought (CoT)~\cite{wei2022chain} reasoning has gained significant attention for its ability to improve model performance by generating intermediate reasoning steps without search. Implicit Chain-of-Thought Reasoning~\cite{deng2023implicit} bypass the need for generating explicit reasoning steps by leveraging the internal hidden states of models. This method distills knowledge from a teacher model trained to generate intermediate steps, allowing student models to solve tasks more efficiently by reasoning vertically through their internal layers. Similarly in chess AI, a 270M parameter transformer model can achieve grandmaster-level play without any explicit search by learning action-values through supervised training on a large dataset of games~\cite{ruoss2024grandmaster}. These approaches highlight a trend where models are increasingly able to generalize complex reasoning and decision-making processes internally, thus reducing reliance on computationally expensive search algorithms while maintaining high performance in domains like mathematical reasoning and game playing.

\paragraph{Self-improvement in LLM}
Self-improvement methods for LLMs aim to enhance model performance by enabling them to learn from their own outputs with minimal human intervention. These approaches typically involve supervised fine-tuning (SFT) on high-quality outputs generated by the models~\cite{10.5555/3600270.3601396,DBLP:conf/iclr/LiYZSLZWL24,wang2024selftaughtevaluators} or preference optimization, where the model learns from pairs of good and bad responses it generated to a query~\cite{xu2024thingscringeothersiterative, DBLP:conf/icml/YuanPCLSXW24, DBLP:journals/corr/abs-2404-19733, DBLP:journals/corr/abs-2407-19594}. In general instruction-following tasks, the quality of model outputs is often determined by an external reward system—this can be a trained reward model~\cite{xu2024thingscringeothersiterative}, human evaluators~\cite{DBLP:journals/corr/abs-1909-08593}, or the LLMs themselves through techniques like LLM-as-a-Judge prompting~\cite{DBLP:conf/nips/ZhengC00WZL0LXZ23}. In mathematical domains, however, output quality is primarily judged by whether the model reaches the correct answer~\cite{10.5555/3600270.3601396,DBLP:journals/corr/abs-2404-19733}. For more fine-grained evaluation, step-level rewards for mathematical reasoning tasks may be assigned by human annotators or a trained process reward model~\cite{DBLP:conf/iclr/LightmanKBEBLLS24}. Iterative self-improvement techniques have shown promise across a range of tasks, from instruction-following~\cite{xu2024thingscringeothersiterative, DBLP:conf/icml/YuanPCLSXW24} to more complex reasoning-based challenges~\cite{10.5555/3600270.3601396, DBLP:journals/corr/abs-2404-19733}, highlighting their potential for driving further advancements in LLM capabilities. However, recent findings suggest that LLM-generated texts often exhibit truncated ``tails'', meaning that the distribution of generated outputs lacks the variability found in human-generated content, particularly in the less common, outlier responses (or "tails" of the distribution)~\cite{shumailov2024curserecursiontraininggenerated, dohmatob2024taletailsmodelcollapse}. This reduced variability can lead to a phenomenon known as model collapse, where the model converges toward a narrower range of behaviors, ultimately harming performance~\cite{shumailov2024curserecursiontraininggenerated}. This issue has been observed in tasks like language modeling~\cite{shumailov2024curserecursiontraininggenerated} and iterative preference optimization for mathematical reasoning~\cite{DBLP:journals/corr/abs-2407-05013}. To mitigate the risk of model collapse, researchers recommend maintaining a balanced mix of clean, human-authored data alongside LLM-generated content during training~\cite{shumailov2024curserecursiontraininggenerated, dohmatob2024taletailsmodelcollapse, gerstgrasser2024modelcollapseinevitablebreaking}. This approach helps preserve diversity and prevents the model from degrading in performance over time.



\section{Exploration Journey}

This section represents the core of our O1 replication endeavor. This section systematically unfolds our exploration process through a series of pivotal questions, mirroring the complex pathway illustrated in our research timeline diagram. From the initial evaluation of O1 using OlympicArena~\cite{huang2024olympicarenabenchmarkingmultidisciplinecognitive} datasets to the intricate ``Long Thought Construction'' phase, our journey has been marked by multiple attempts, continuous iterations, and deep dives into the essence of O1's capabilities.

The questions we address in this chapter not only reflect the progression of our research but also embody our profound inquiry into the nature of O1's cognitive processes. 
We begin by examining the structure of O1's thoughts, then delve into the mechanics and construction of long thoughts - a concept central to our ``Long Thought Construction'' phase depicted in the diagram. Our exploration extends to the development of reward models, the construction of on-policy reasoning trees, and the integration of these elements into cohesive long thoughts, mirroring the complex web of interconnected processes in our research timeline.
Our methodology, as visualized in the diagram, involves multiple iterations and parallel streams of investigation. This approach is reflected in our discussion of evaluation methods and training strategies, showcasing how we validate hypotheses and refine our techniques through cycles of quantitative and qualitative assessments, including human checks and specialized analysis tools.

By structuring this chapter around these key questions, we not only provide a clear narrative of our technical journey but also demonstrate a systematic approach to exploring unknown AI technologies. This question-driven format aligns with our ``journey learning'' paradigm, emphasizing the importance of the entire learning and exploration process, not just the final outcomes.
As we progress through each question, readers will gain insights into our decision-making process, the challenges we faced, and the innovative solutions we developed. This transparent sharing of our thought processes, attempts, and even failures, as illustrated in our research timeline, aims to contribute valuable insights to the AI community and foster collective advancement in the field.

Through this section, we invite readers to traverse our exploration journey, understanding not just what we discovered about O1, but how we approached the daunting task of replicating a groundbreaking AI model with limited information. Our journey, marked by curiosity, persistence, and innovation, serves as a testament to the power of open, collaborative AI research in pushing the boundaries of what's possible in artificial intelligence.


\subsection{Q1: What does O1's Thought Look Like?} \label{subsec:looklike}
The Table~\ref{tab:example_statistics} is created based on a detailed analysis of O1's thought examples provided by OpenAI,~\footnote{\url{https://openai.com/index/learning-to-reason-with-llms/}} which includes eight instances of reasoning steps, or ``thoughts,'' for solving complex tasks. Each example in this is meticulously examined to extract relevant features such as the number of tokens, lines, and keyword.
These examples are categorized into different problem types, each associated with a difficulty level ranging from simple English reading comprehension to complex multi-step math reasoning tasks. Our analysis demonstrates a trend: \textbf{as the difficulty increases, the response length (both tokens and lines) tends to grow proportionally. This suggests that higher difficulty problems involve more reasoning steps.}

In addition to token and line counts, we conducted a keyword frequency analysis to identify recurring terms that may characterize the reasoning process. In addition to commonly observed connective words like ``and'' and ``so'', our analysis highlights several less frequently occurring but highly significant keywords. Keywords such as “consider”, “if” and “possible” appear frequently, \textbf{often signaling branching in the reasoning process where multiple paths are considered}. The frequency of these keywords was notably higher in problems with higher complexity, indicating the model's exploration of different solution paths in these scenarios. Keywords like \textbf{``wait'' and ``Alternatively'' are crucial indicators of the model's ability to engage in reflection and self-correction}. This suggests a deeper understanding and a more nuanced approach to reasoning, as the model is not just following a linear path but is capable of reconsidering and refining its approach based on reflection.

To understand the thought process of OpenAI's O1, we consult two PhD candidates from the mathematics department to carefully review the reasoning process employed by OpenAI's O1 in solving mathematical problems. Through their detailed examination, they extracted the underlying thought chain that reflects how O1 approaches and reasons through complex equations. This structured thought graph is illustrated in Figure~\ref{fig:o1-math-graph}.
After these explorations, we determined that the long thought data we need to construct should have the following characteristics:

\begin{itemize}
    \item Iterative Problem-Solving: The model starts by defining functions and gradually explores related expressions, breaking down complex equations into simpler components, reflecting a structured and methodical approach.
    \item Key Thought Indicators: The use of terms like ``Therefore'' for conclusions, ``Alternatively'' for exploring different paths, ``Wait'' for reflection, and ``Let me compute'' for transitioning into calculations highlights the model's reasoning stages.
    \item Recursive and Reflective Approach: The model frequently reassesses and validates intermediate results, using a recursive structure to ensure consistency, which is typical in rigorous mathematical reasoning.
    \item Exploration of Hypotheses: The model tests different hypotheses, adjusting its approach as it gathers more information, demonstrating flexibility in its reasoning process.
    \item Conclusion and Verification: Finally, the model solves the equations and verifies the results, emphasizing the importance of validating conclusions before finishing.
\end{itemize}






\begin{table*}[!t]
    \centering
    \setlength{\tabcolsep}{6pt}
    {
    \fontsize{10}{13}\selectfont 
    \renewcommand\arraystretch{1.5}
    \resizebox{\linewidth}{!}{
        \begin{tabular}[l]{p{0.15\linewidth}p{0.08\linewidth}p{0.08\linewidth}p{0.11\linewidth}p{0.40\linewidth}}
        \toprule
        \textbf{Example} & \textbf{Token Count} & \textbf{Line Count} & \textbf{Avg. Words per Line} & \textbf{Keyword Count} \\ \midrule
        Cipher & 8915 & 668 & 4.29 & So: 31, First: 27, of: 27, Alternatively: 21, Second: 19, Third: 15, But: 15, Wait: 13, Alternatively perhaps: 13, Let: 12, and: 12, let: 11, first: 9, Now: 9, Think step: 8, step by: 8, by step: 8, So the: 8, the first: 8, Similarly: 6, as: 6, or: 5 \\ \midrule
        Coding & 3259 & 197 & 3.64 & and: 8, as: 7, then: 4, For: 4, Now: 4, So: 3, Let: 3, Since: 3, We: 3, row=: 3, step by: 2, by step: 2, We need: 2, Let me: 2, step by step: 2, We need to: 2, Let me try: 2 \\ \midrule
        Crossword & 5311 & 396 & 5.75 & Across: 37, So: 33, From: 31, and: 25, first: 19, Position: 19, we: 15, Now: 13, Possible: 13, as: 7, But: 7, Similarly: 7, third: 7, First: 6, So we: 6, Given: 5, Now let: 5, We: 4 \\ \midrule
        English & 757 & 49 & 9.88 & the: 20, that: 15, to: 15, is: 13, because: 8, why: 7, Option because: 5, and: 4 \\ \midrule
        Health Science & 1010 & 86 & 6.14 & and: 11, So: 5, Also: 3, But: 3, First: 2, Then: 2, So the: 2, but also: 2 \\ \midrule
        Math & 18751 & 521 & 9.49 & Therefore: 48, But: 42, So: 38, Thus: 36, Similarly: 33, we: 26, and: 17, since: 16, for: 15, real: 15, Wait: 14, Let: 10, but: 9, Let me: 9, all: 8, k1: 8, Given: 8, Wait but: 8, Alternatively: 7, we can: 7, So the: 7, Then: 6, Given that: 6 \\ \midrule
        Safety & 510 & 41 & 8.27 & So: 6, and: 65, But: 3, Also: 2, ChatGPT: 1, Write: 1, Explain: 1 \\ \midrule
        Science & 2411 & 91 & 7.62 & and: 14, can: 6, compute: 6, But: 5, So: 4, Now: 3, Given: 2, but: 2, so: 2, Alternatively: 2 \\ 
        \bottomrule
        \end{tabular}
    }
    \caption{Statistical summary of various examples from OpenAI O1’s thought process across different domains. The table presents key metrics including the token count, the line count, the average number of words per line, and the frequency of the highest-occurring words or phrases derived using the n-gram algorithm. These keywords reflect the structure and style of the reasoning process, highlighting how the model introduces logical steps, alternatives, or corrections in different contexts.}
    \label{tab:example_statistics}
    }
\end{table*}

\definecolor{wkblue}{rgb}{0.2, 0.3, 0.6}
\definecolor{meta-color}{rgb}{0.5, 0.5, 0.5}

\begin{figure}
    \centering
    \input{o1-math-graph}
    \caption{The Thought Structure of OpenAI O1 in Mathematical Reasoning.}
    \label{fig:o1-math-graph}
\end{figure}

\begin{figure}
    \centering
    \input{our-case2}
    \caption{Generated Thought by Our Proposed Model in Mathematical Reasoning.}
    \label{fig:our-math-case2}
\end{figure}

\begin{figure}
    \centering
    \input{our-case1}
    \caption{Generated Thought by Our Proposed Model in Mathematical Reasoning.}
    \label{fig:our-math-case1}
\end{figure}

\begin{figure}
    \centering
    \input{our-case3}
    \caption{Generated Thought by Our Proposed Model in Mathematical Reasoning.}
    \label{fig:our-math-case3}
\end{figure}

\subsection{Q2: How does Long Thought Work?}
This is a question we consider important. However, at our current stage of progress, we are merely putting forward our hypotheses. We don't believe we have sufficient empirical evidence to verify their accuracy.
The remarkable success of O1's long-thought approach can be attributed to journey learning, which we have introduced in \S\ref{sec:journey-learning}.
Unlike traditional shortcut learning, journey learning allows the model to explore the \textbf{entire decision trajectory}, mimicking human problem-solving processes. This comprehensive exploration enables O1 to consider \textbf{multiple solution paths, learn from errors, and understand the complete problem-solving process}. By experiencing both correct and incorrect paths, the model develops robust error-handling and self-correction capabilities, enhancing its adaptability to new challenges. This approach fosters a deeper understanding of the problem domain, \textbf{going beyond merely knowing the correct answer to comprehending why and how to arrive at it}. The journey learning process closely simulates human cognitive processes, incorporating trial-and-error, reflection, and adjustment. This results in enhanced explainability, as O1 can provide detailed solution steps and explain its reasoning, including how it recovers from mistakes. Consequently, O1's long thought process, grounded in journey learning, is not simply about extended computation time but represents a thorough, human-like reasoning exploration. This methodology equips O1 to handle more complex problems, offer more reliable and interpretable answers, and demonstrate greater adaptability when faced with novel challenges, thus explaining its exceptional performance across various tasks.



\subsection{Q3: How to Construct Long Thoughts?}
Constructing long thoughts with actions such as reflection and backtracking is the core part of journey learning.
To achieve this, we undertook a series of attempts.


\paragraph{Attempt 1: Tree Search with LLM and Reward}
Based on our observations of long thought in \S\ref{subsec:looklike}, its most prominent feature is the attempt to reflect and backtrack when reasoning leads to an incorrect or unhelpful node. This resembles searching on a reasoning tree for a problem, backtracking at erroneous nodes, until the correct solution path is found. To achieve this, we need to construct a reasoning tree where the root node represents the problem, and each other node represents a reasoning step. The path from the root to any node signifies the reasoning process from the problem to that conclusion. Moreover, backtracking and reflection must be based on incorrect reasoning steps, necessitating a more fine-grained reward model (i.e., process-level) to indicate the correctness of each node in the tree. By executing a search algorithm on a reasoning tree with process-level rewards, we can integrate erroneous steps into a chain of thought, thereby constructing long thought that encompasses actions like backtracking and reflection.

\paragraph{Attempt 2: Propose-Critique Loop}
Attempt 1 constructs long thought by executing searches on the tree based on predefined rules, but this limits the freedom of actions like backtracking and reflection. Therefore, we allow the model to choose its current actions. We constructed a Propose-Critique Loop, where we pre-define some possible actions for the model (i.e., continue, backtracking, reflection, terminate) and let the model select actions to build the reasoning tree. If the tree does not reach the final answer, the model can be informed of this negative signal, guiding it to reflect and correct its approach.

\paragraph{Attempt 3: Multi-Agent Approach}
Building long thought on the foundation of a reasoning tree presents several challenges, including the presence of numerous ineffective nodes that do not contribute to constructing Long Thought, as well as issues of logical inconsistency caused by reasoning steps that do not depend on the reflection behavior. To address this, we designed an algorithm utilizing multi-agent debate, where one agent acts as the policy model, continuously reasoning, while another agent serves as the critique model, indicating whether the policy model should continue with the current reasoning or perform actions like backtracking. The two agents engage in ongoing dialogue, naturally constructing a long thought dataset when the correct answer is found.

\paragraph{Attempt 4: Complete Human Thought Process Annotation}
When humans tackle reasoning problems, they typically do not engage in constant forward reasoning until they either solve the problem or fail; instead, they reflect, backtrack, and rewrite reasoning when they can no longer proceed. This behavior closely aligns with the characteristics of long thought. Thus, we can faithfully and comprehensively document the process by which humans solve reasoning tasks, resulting in high-quality long thought.

\subsection{Q4: How to Construct Reward Models?}

The first step in utilizing the reward model is to define the granularity. Instead of focusing solely on the final results, we aim to enhance the capabilities of LLMs specifically in reflection, backtracking, and related cognitive processes. Therefore, we define the evaluation granularity at the step level. Specifically, we use the fine-tuning data from~\citet{abel} to make the solutions distinct by line numbers.
The process of implementing the reward model can involve using either open-source reward models or proprietary models. We compare the performance of different reward models on the subsets of PRM800K~\citep{DBLP:conf/iclr/LightmanKBEBLLS24} and MR-GSM8K~\citep{DBLP:journals/corr/abs-2312-17080}. We present the results in Table~\ref{tab:MR-GSM8K} and Table~\ref{tab:PRM800K}. O1-mini performs best across different datasets.

\begin{table}[htbp]
  \centering
  \begin{minipage}{0.45\linewidth}
    \centering
    \begin{tabular}{lc}
    \toprule
    Model & F1 score \\
    \midrule
    o1-mini & \textbf{0.855} \\
    \midrule
    GPT-4o-mini & 0.722 \\
    \midrule
    Math-shepherd & 0.734 \\
    \midrule
    ReasonEval-7B & 0.728 \\
    \midrule
    ReasonEval-34B & 0.735 \\
    \bottomrule
    \end{tabular}%
        \caption{Results on the subset of MR-GSM8K}
    \label{tab:MR-GSM8K}%
  \end{minipage}
  \begin{minipage}{0.45\linewidth}
    \centering
    \begin{tabular}{lc}
    \toprule
    Model & F1 Score \\
    \midrule
    GPT-4o-mini & 0.756 \\
    \midrule
    o1-mini & \textbf{0.880} \\
    \midrule
    o1-preview & 0.867 \\
    \bottomrule
    \end{tabular}%
        \caption{Results on the subset of PRM800K}
    \label{tab:PRM800K}%
  \end{minipage}
\end{table}

\subsection{Q5: How to Construct an On-policy Reasoning Tree?}
The construction of a reasoning tree requires a policy model \(\pi\) that can perform single-step reasoning. Given a problem \(q\) and its corresponding final answer \(a\), \(\pi\) starts from the problem as the root node and continuously adds new nodes to the tree. It first generates \(w\) possible first-step reasoning steps as child nodes of the root node. Then, it iteratively performs forward reasoning, generating \(w\) possible subsequent reasoning steps for each current node (e.g., the first-step reasoning) as child nodes of that node. This process is repeated until a preset maximum depth \(D\) is reached or all leave nodes reach the final answer.

\paragraph{Policy Model and Step Segmentation}  
Constructing the reasoning tree requires a clear definition of reasoning steps. To this end, we adopt the data format proposed in Abel~\citep{abel}, transforming mathematical problem solutions into a form with clear steps, dividing answers into multiple lines, each beginning with a line number and including reasoning within the line. Thus, we fine-tuned DeepSeekMath-7B-Base~\citep{shao2024deepseekmath} using the dataset from Abel to obtain Abel-DSMath, serving as the policy model \(\pi\). The model fine-tuned on this specific format data can conveniently control the generation of individual reasoning steps.

\paragraph{Reward Model and Pruning}  
The tree generation algorithm proposed above is computationally expensive. When setting \(w\) to 3 and \(D\) to 10, the last iteration requires generating \(3^{10}\) reasoning steps. Therefore, we use a reward model to prune erroneous reasoning steps, improving operational efficiency. Specifically, we employ beam search, selecting only a small number of candidates for retention in each iteration for the next round. Depending on the reward model used, the details of pruning implementation vary. We attempted two reward models: math-shepherd~\cite{wang2024mathshepherdverifyreinforcellms} and o1-mini. 
Math-shepherd provides a real number between 0 and 1 for each step, representing the probability of correctness for the current step. In each iteration of tree generation, we score all reasoning steps and select the top \(K\) with the highest scores for the next iteration. This reduces the total generation count from \(\frac{n^D-1}{n-1}\) to \(nKD\). However, math-shepherd struggles to effectively evaluate reasoning steps for difficult problems, necessitating a more robust reward model that offers high accuracy in correctness indications for each step. Thus, finally, we use o1-mini to provide rewards for each step, directly indicating whether each reasoning step is correct or incorrect. At this point, in each iteration of tree generation, we utilize the rewards from o1-mini and select at most \(K\) correct reasoning steps for the next iteration.

\subsection{Q6: How to Derive a Long Thought from a Reasoning Tree?}
Once the reasoning tree is constructed, our goal is to derive a long thought from the tree that incorporates trial and error. This approach contrasts with traditional methods that focus solely on a shortcut to the correct answer and valid intermediate steps. In our framework, each node of the reasoning tree is annotated with a rating from a reward model that indicates whether the step is correct or incorrect, along with reasoning that justifies this judgment.

\paragraph{Constructing the ShortCut from a Reasoning Tree}
We first construct the shortCut from the reasoning tree, which includes only the correct answer and valid intermediate steps. Starting from the root node, which represents a question, we identify a path that leads to a correct answer leaf node. If there are multiple correct answer nodes, multiple correct paths will be established.

\paragraph{Traversal Path from a Reasoning Tree} 
To derive a long thought, we employ a Depth First Search (DFS) traversal of the tree. This traversal constructs a path in DFS order, documenting each step from the root question node to a correct answer leaf node while including reasoning for any node marked as incorrect. The challenge with DFS lies in its exploration of a vast search space, resulting in numerous trial-and-error paths that may not yield a correct solution. To simplify this initial exploration, we introduce specific constraints to manage the complexity.

Initially, we mark all nodes in the tree based on whether they lie on the correct path (i.e., the shortCut). The traversal adheres to the following rules:
(i). Nodes on the correct path: We allow exploration of child nodes that are not on the correct path. This means that when DFS encounters a node on the correct path, it may explore a child node that leads to an incorrect outcome. Once this node reaches a leaf node and is determined to be incorrect, the algorithm backtracks to continue traversing along the correct path.
(ii). Nodes not on the correct path: The traversal randomly selects one child node to explore without branching into trial and error.
To further streamline the process, we apply an additional constraint: each node on the correct path is permitted a maximum of K trials—one trial on an incorrect path and one on the correct path.

These constraints ensure that the DFS traversal focuses on a manageable subset of the search space, allowing for meaningful trial-and-error exploration while avoiding excessive exploration of incorrect paths. In future experiments, we plan to remove or adjust these constraints to investigate the relationship between the length of trial paths and the performance of the final model.

\paragraph{Long Thought from a Traverse Path} 
With the traversal path generated and reasoning attached to the wrong nodes, we construct a draft long thought by concatenating all steps in the path. This draft incorporates the reasoning for each incorrect step. However, initial experiments using this raw draft to train models have demonstrated suboptimal performance.
To address this, we employ GPT-4o to modify the draft. GPT-4o enhances the coherence and smoothness of the thought process while preserving all reasoning steps, including incorrect steps, reflections, and corrections. This approach ensures that the final long thought is not only accurate but also flows naturally, simulating the human problem-solving process with both correct and incorrect steps.


\subsection{Q7: How to Evaluate our Trials?}
In addition to testing accuracy scores using specific evaluation metrics on benchmarks, manually reviewing actual cases is a crucial step in evaluating data and models. Therefore, to provide a more intuitive way to evaluate the model’s performance on specific problems, we build a visual data analysis platform using Streamlit.~\footnote{https://streamlit.io/}
Specifically, our visualization platform includes the visualization of synthetic trees and their corresponding long thoughts as well as the output of the trained model. Furthermore, when visualizing results, we support detailed conditional filtering, such as filtering for correctly or incorrectly answered questions, or whether the output contains keywords indicating reflection or hesitation (e.g., “wait”). Additionally, we support comparison between different iterations of synthetic data and model outputs, which makes it highly intuitive and helps us easily validate whether the new round of data or models is effective.


\subsection{Q8: How to Train our Models?}
Our experiments utilize the pre-trained language model deepseek-math-7b-base.~\footnote{More other models have already been in our waiting list.} The training process is divided into two main phases: Supervised Fine-Tuning (SFT) and Direct Preference Learning (DPO)~\citep{rafailov2024direct}.

\paragraph{Phase 1: Supervised Fine-Tuning (SFT)}
The SFT process consists of two stages: 1. \textbf{ShortCut Learning:} In this initial stage, we focus on fine-tuning the model using responses that include only the correct intermediate steps and the final correct answer. We fine-tune Deepseek-math-7b-base~\citep{shao2024deepseekmath} on the Abel dataset~\citep{abel}, which comprises 120k examples, and the PRM800K dataset~\citep{DBLP:conf/iclr/LightmanKBEBLLS24}. For each question in PRM800K, we utilize a single correct step-by-step solution, discarding responses that do not lead to the final answer. This results in a total of 6,998 examples for fine-tuning. During this stage, we conduct fine-tuning for one epoch on each dataset, primarily aiming to familiarize the model with the desired response format. 2. \textbf{Journey Learning:} In this second stage, we further fine-tune the initial stage SFT model using the long thoughts we constructed, which comprise 327 examples. This phase is designed to enhance the model's ability to detect errors, incorporate reflections, execute corrections, and perform backtracking. By training on long thoughts that include not only the correct reasoning paths but also erroneous trials, we aim to equip the model with a deeper understanding of the complexities involved in longer reasoning chains. As a comparison, we also fine-tune the model on the corresponding shortCut generated from the same reasoning tree, which also consists of 327 examples. Both the long thought SFT and shortCut SFT settings are trained for 3 epochs on these 327 examples.

\paragraph{Phase 2: Direct Preference Learning (DPO)}
In this phase, we generate 20 responses per question from the MATH Train dataset, a re-divided dataset from PRM800k that includes 12,000 examples, using nucleus sampling with $ \text{top}\_p = 0.95 $ and temperature $ T = 0.7 $. These 20 responses are categorized into positive and negative responses based on the correctness of the final answer. From these, we randomly select 5 positive responses and 5 negative responses to create 5 preference pairs. We then train the model using these preference pairs with DPO loss, allowing it to learn from the comparison of correct and incorrect answers.

The results of our experiments are shown in Table \ref{tab:train_result}. All results are tested on the MATH test set, using a re-divided subset from PRM800K, which includes 500 examples. The results show that Journey Learning led to significant improvements compared to Shortcut Learning, with gains of +8.4 and +8.0 on the deepseek-sft-abel and deepseek-sft-prm800k models, respectively, demonstrating the effectiveness of our proposed Journey Learning method. However, the improvement from DPO was more modest, and we acknowledge that this is an initial exploratory result. In future experiments, we plan to further explore preference learning and Reinforcement Learning (RL) techniques. This will include, but not be limited to, iterative self-improvement, incorporating process-level reward models, and transitioning from outcome-level DPO to process-level DPO/RL approaches.

\begin{table}[!htp]\centering
\begin{tabular}{lrrr}\toprule
&deepseek-sft-abel &deepseek-sft-prm800k \\\midrule
SFT-phase1 &0.372 &0.290 \\
SFT-phase2-shortcutLearning &0.386 &0.348 \\
SFT-phase2-journeyLearining &0.470 &0.428 \\
DPO &0.472 &0.440 \\
\bottomrule
\end{tabular}
\caption{Training Results on MATH Test Set}\label{tab:train_result}
\end{table}

\subsection{Q9: What Would be an Effective Annotation Strategy for Human-AI Collaboration?} 

    
We have developed a human-AI pipeline designed to generate high-quality, long-form reasoning data for problems derived from the MATH dataset. This pipeline enables the expansion of a human-annotated solution of several lines into thousands of tokens, which follows our “journey learning” paradigm. During the pipeline’s construction, we identified key techniques for efficient annotation, including:

\paragraph{Complete Thought Process} It is not essential for annotators to record every word that comes to mind in detail, but \textbf{it is crucial to document each trial, reflection, association, and correction}. These diverging cognitive pathways may not always be explicitly expressed or consciously recognized in everyday thinking. Nevertheless, capturing shifts in thought, along with the reasons behind these shifts, is critical. This ability to navigate and understand cognitive transitions is a core skill that large language models must learn from our data.

\paragraph{Additional Explanation for Common Sense} Humans often omit information that can be inferred from context, such as references to previously mentioned formulas or the application of well-known theories. However, this can lead to hallucination when large language models attempt to interpret human annotations. Therefore, high-quality data must include explicit explanations of common-sense knowledge to prevent misinterpretation by LLMs.

With the essential components outlined previously, the concise yet precise annotated data is fully generated by human effort. The next stage involves AI-driven processes. By designing sophisticated prompts, we implement data augmentation by LLMs in aspects below:

\begin{enumerate}
    \item \textbf{Enhancement of Data Granularity} The prompt emphasizes breaking down the problem-solving process into finer, smaller steps. By splitting the process into fine-grained, easily digestible chunks, it becomes easier for LLMs to grasp and internalize each concept before moving on to the next. This ensures deeper comprehension at every stage.
    \item \textbf{Gradual Reasoning} LLMs are required to frequent pause, reflect on known information or to clarify the next step should be added to help guide reasoning. Taking pauses in reasoning mimics how students would naturally think about the problem, helping them stay engaged and connected to the reasoning process rather than passively following instructions.
    \item \textbf{Student-Explorer Perspective} Instead of presenting the solution as if the answer is already known, LLMs are encouraged using a tone of discovery, where the they solving the problem is thinking through it for the first time. This fosters curiosity and encourages students to think critically, making them feel like they are part of the learning process rather than simply receiving information.
\end{enumerate}



\section{Detailed Event Explanation of Our Research Exploration}

\newpage
{\tiny
\renewcommand{\arraystretch}{1.45}
\begin{longtable}{@{}lp{3.8cm}p{7cm}p{3cm}@{}}
\toprule

        Node & Short Description & Explanation & Resource \\ \midrule

        1 & OpenAI o1 Release & OpenAI released its latest reasoning model, o1 &  \\ \midrule 
        2 & Evaluate o1 (OlympicArena, Gaokao Math) & Evaluate the performance of o1 on high-difficulty competition questions & OlympicArena, Gaokao,\newline o1 API \\ \midrule
        3 & Knowledge Acquisition & Learn about the possible technical routes for OpenAI o1 &  \\ 
        4 & 1st o1 Technical Discussion & Discuss the technical route of o1 and determine research goals &  \\ 
        5 & Team Assembled & Gather relevant students to form a team &  \\ \midrule
        6 & o1 Thought Analysis \& Schema Design & 1. Analyze the properties/schema of o1 long thought: long thought structure, functions of each part \newline 2. Explore how to construct long thought training data: Use the MATH dataset & OpenAI o1 Examples \\ \midrule
        7 & Attempt1: Propose-Critique Loop & Multi-agent system where Propose suggests possible reasoning steps and Critique points out issues and suggests directions & Proposer, Critic, Loop Algorithm \\ 
        8 & Attempt2: Tree Search with LLM and Reward & Build a reasoning tree for a reasoning problem using PolicyModel and RewardModel, where each node represents a step, and use the reasoning tree to construct long thought & Policy Model, Reward Model, long thought Construction Algorithm \\ 
        9 & Attempt3: Multi-Agent Approach & Use a multi-agent debate system to solve reasoning problems and integrate reasoning paths, including reflection and backtracking, into long thought data & Policy Model, Reward Model, Algorithm \\ 
        10 & Attempt4: Complete Human Thought Process Annotation & Human experts create a small amount of high-quality long thought data & Human \\ \midrule
        11 & Process-level Reward Model & Used to score each reasoning step in the reasoning tree, providing reasons & Reward Model \\ 
        12 & Construct of Reasoning Tree & Construct a reasoning tree where each node represents a reasoning step & Policy Model, Search Algorithm \\ 
        13 & Integrating Reasoning Tree into Long Thought & Use the reasoning tree to construct long thought data that includes backtracking and reflection, rather than a direct forward reasoning chain & Reasoning Tree; Long Thought Construction Algorithm \\ 
        14 & Teacher-Student Incentive-Driven Data Construction & Student (Policy Model) continuously reasons forward, while Teacher (Critique Model) provides feedback to Student, points out errors, and helps with backtracking and reflection & Policy Model, Reward Model, Algorithm \\ 
        15 & Reward \& Critique Setup & The selection of Reward Model or Critique Model includes using the open-source model MathShepherd to score steps 0-1, or using powerful closed-source models like o1-mini to directly indicate the correctness of steps & MathShepherd, o1-mini \\ 
        16 & On-Policy Sampling \& Searching Tree & On-policy setting, using the target model to be improved as the policy model to provide reasoning steps. To speed up tree construction, use the reward model and corresponding algorithms to prune the tree during its construction & Policy Model, Reward Model, Math Training Set \\ 
        17 & Off-Policy PRM800k Tree & OpenAI officially released PRM800k, a dataset that contains both Reasoning Tree and Process-Level Reward, used to construct the corresponding Reasoning Tree with reward. Since reasoning steps are provided by humans rather than the target model to be improved, it is set as off-policy & PRM800k Dataset \\ 
        18 & 1st Construction of Reasoning Tree & Construction of the first-generation reasoning tree. Train the Policy Model M1 using DeepSeekMath-Base-7B on Abel data and use MathShepherd to provide the Process Level Reward Model. Generate and prune the tree using the Beam Search algorithm & Policy Model: DeepSeekMath-7B + Abel; Reward Model: Mathshepherd-Mistral-7B \\ 
        19 & 1st Round of Long Thought Integration & Use PRM800k data to synthesize long thought training data & 1st Reasoning Tree; long thought Construction Algorithm \\ \midrule
        20 & Evaluation & Evaluate the synthesized long thought & Evaluation Methods, Long Thought Data \\ 
        21 & Training & Train models using long thought & Long Thought, Model \\ 
        22 & Pretrain & Pre-train the model using large amounts of data & Massive Long Thought Data, Model \\ 
        23 & Post-train & Fine-tune the model using long thought on the pre-trained model & Long Thought Data, Pretrained Model \\ 
        24 & Iterative Train & Iterative training to improve the model & Fine-tuned Model \\ 
        25 & Preference Learning & Preference learning, enabling models with reflection and backtracking capabilities to automatically select more effective answering strategies & SFT Model; Data \\ 
        26 & SFT & Supervised learning, directly training the model with long thought data & Pretrained Model; Long Thought Data \\ 
        27 & RL & Reinforcement learning, such as PPO & SFT Model; Reward Model \\ 
        28 & DPO & Direct preference optimization, a stable preference learning algorithm & Preference data; SFT Model \\ 
        29 & Analysis Tool & A platform for visualizing long thought data and analyzing model responses & Streamlit \\ 
        30 & Human Check & Human experts analyze and evaluate long thought data and model responses & Human; Long Thought Data; Model Response \\ \midrule
        31 & 2nd Round of Long Thought Integration & Use the constructed second-generation reasoning tree to synthesize long thought training data & 2nd Reasoning Tree; long thought Construction Algorithm \\ 
        32 & 2nd Construction of Reasoning Tree & Build the second-generation reasoning tree, replacing the Reward Model with o1-mini, which directly indicates the correctness of steps and performs pruning, providing more accurate rewards & Policy Model: DeepSeekMath-7B + Abel; Reward Model: o1-mini \\ \midrule
        33 & Fine-Grained, Thought-Centric Evaluation & Conduct a more fine-grained evaluation of synthesized long thought data to enhance the effectiveness of various actions within long thoughtt & Long Thought Data \\ 
        34 & Experiments on Long Thought Scaling Law & Experiment with the scaling law of long thought in terms of training time and inference time & Massive Long Thought Data of Various Formation \\ 
        35 & Human-AI Collaboration for Quality Thought & Use human-generated high-quality long thought data & Human \\ 
        36 & 3rd Round Long Thought Integration & Synthesize the third-generation long thought data, further improving quantity and quality & Massive Reasoning Trees; Construction Algorithm \\ 
\bottomrule
\caption{Detailed event explanation of our research exploration. The nodes and short descriptions correspond to node in Figure ~\ref{research_journey}, while the explanations and resources represent a detailed elaboration of the purpose of the node and the relevant resources required.} 
\label{details_table}
\end{longtable}}



\section{Future Plan}

As our O1 Replication Journey continues to evolve, our future plans are shaped by the insights gained and challenges encountered thus far. Drawing from our research timeline and the progress we've made, we've identified several key areas for future exploration and development:

\begin{enumerate}
    \item \textbf{Scaling Up Long Thought Integration}:
Building on our successful iterations of long thought integration, we plan to conduct a third round of integration, as indicated in our research diagram. This will involve scaling up our processes to handle more complex and diverse thought patterns, potentially uncovering new dimensions of O1's capabilities.
    \item \textbf{Experiments on Long Thought Scaling Laws}:
Our diagram highlights planned experiments on long thought scaling laws. This research stream aims to understand how the performance and capabilities of our model scale with increases in data, model size, and computational resources. These insights will be crucial for optimizing our approach and potentially discovering fundamental principles underlying advanced AI systems.
    \item \textbf{Fine-Grained, Thought-Centric Evaluation}:
We plan to develop and implement more sophisticated evaluation methodologies, focusing on fine-grained, thought-centric assessment. This approach, highlighted in our research timeline, will allow us to more accurately measure the quality and coherence of the generated long thoughts, providing deeper insights into our model's reasoning capabilities.
    \item \textbf{Human-AI Collaboration for Quality Thought}:
A key component of our future plan, as shown in the diagram, is to explore and enhance human-AI collaboration for producing high-quality thoughts. This involves developing interfaces and methodologies that leverage the strengths of both human intelligence and AI capabilities, potentially leading to breakthroughs in hybrid intelligence systems.
    \item \textbf{Continued Improvement of Reward and Critique Models}:
Building on our process-level reward model and critique model setup, we aim to refine these systems further. This ongoing process will involve iterative improvements to better capture the nuances of human-like reasoning and problem-solving strategies.
    \item \textbf{Advanced Integration of Reasoning Trees}:
We plan to explore more sophisticated methods of deriving and integrating long thoughts from our reasoning trees. This will involve developing advanced algorithms for traversing and synthesizing information from these complex structures.
    \item \textbf{Expansion of Training Methodologies}:
Our future plans include further experimentation with and refinement of our training pipeline. This encompasses enhancements to our pre-training, iterative training, reinforcement learning, preference learning, and DPO (Direct Preference Optimization) stages, as outlined in our research diagram.
    \item \textbf{Continued Transparency and Resource Sharing}:
In line with our commitment to open science, we will continue to share resources, insights, and tools developed throughout our journey. This ongoing practice, represented by the resource-sharing icons in our diagram, aims to foster collaboration and accelerate progress in the wider AI research community.
    \item \textbf{Exploration of Multi-Agent Approaches}:
Building on our initial attempts with multi-agent systems, we plan to delve deeper into this area, potentially uncovering new ways to model complex reasoning and decision-making processes.
    \item \textbf{Refinement of Analysis Tools}:
We aim to further develop and enhance our analysis tools, as indicated in our research timeline. These tools will be crucial for interpreting model outputs, tracking progress, and guiding future research directions.
\end{enumerate}

By pursuing these avenues, we aim to not only advance our understanding and replication of O1's capabilities but also to push the boundaries of AI research methodologies. Our future plans reflect our commitment to the journey learning paradigm, emphasizing continuous improvement, transparent exploration, and collaborative advancement in the field of artificial intelligence.
As we move forward, we remain adaptable to new discoveries and challenges, ready to adjust our plans as our understanding of O1 and advanced AI systems continues to evolve. Through this ongoing journey, we hope to contribute significantly to the development of more capable, interpretable, and ethically aligned AI systems.

\section*{Acknowledgment}
We would like to express our sincere gratitude to the Shanghai Innovation Institute for providing an exceptional environment for our discussion and debate, which has laid a solid foundation for the advancement of our project.
We deeply indebted to all co-authors, with special appreciation extended to the students from GAIR. Your dedication in sacrificing your National Day holiday to collaborate on this groundbreaking endeavor is truly commendable. The progress of this project would not have been possible without your unwavering commitment and hard work.

\end{document}